\documentclass[10pt,twocolumn,letterpaper]{article}

\usepackage[final]{iccv}      
\usepackage{multirow}
\usepackage{makecell}
\usepackage{booktabs}
\usepackage{overpic}
\usepackage{adjustbox}

\definecolor{iccvblue}{rgb}{0.21,0.49,0.74}
\usepackage[pagebackref,breaklinks,colorlinks,allcolors=iccvblue]{hyperref}

\def\paperID{7945} 
\def\confName{ICCV}
\def\confYear{2025}

\title{Stroke-based Cyclic Amplifier: Image Super-Resolution at Arbitrary Ultra-Large Scales
}

\author{Wenhao Guo$^{1}$ \quad 
Peng Lu$^{1}$\thanks{Corresponding author.} \quad 
Xujun Peng$^{3}$ \quad
Zhaoran Zhao$^{1}$ \quad 
Sheng Li$^{2}$ \\
$^{1}$Beijing University of Posts and Telecommunications \\
$^{2}$Peking University \quad 
$^{3}$Amazon AGI Foundations \\
{\tt\small\{whguo,lupeng,zhaoranzhao\}@bupt.edu.cn}\\
{\tt\small penxujun@amazon.com, lisheng@pku.edu.cn}
}

\begin{document}

\twocolumn[{
\renewcommand\twocolumn[1][]{#1}
\maketitle
\begin{center}
    \captionsetup{type=figure}
    \begin{minipage}[h]{0.95\linewidth}
        \centering
        \begin{minipage}[h]{1\linewidth}
            \centering
            \includegraphics[width=1\linewidth]{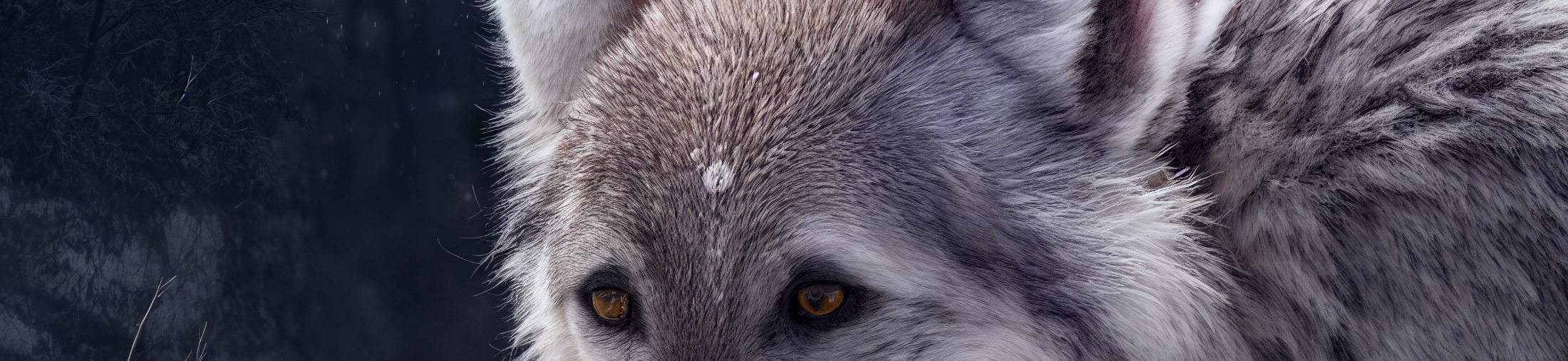}
        \end{minipage}\\
        \vspace{0.001\linewidth}
        \begin{minipage}[h]{1\linewidth}
            \centering
            \includegraphics[width=1\linewidth]{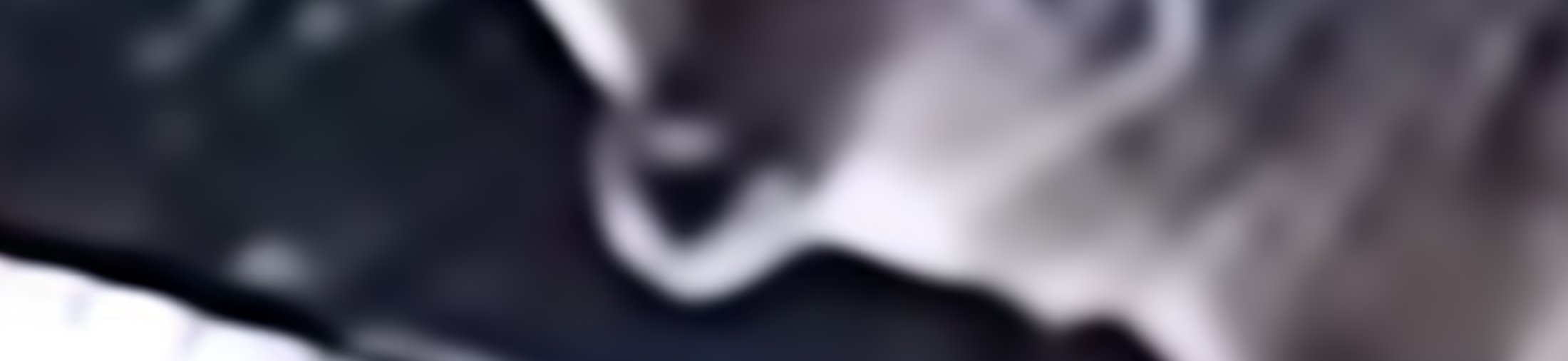}
        \end{minipage}
    \end{minipage}
    \captionof{figure}{
    Comparison of ×100 super-resolution using benchmark dataset BSD100 (img-29): Our SbCA (top) compared to the state-of-the-art ASSR method CiaoSR\cite{ciaosr} (bottom). For more visual results, please refer to the supplementary video.}\label{fig:large} 
\end{center}
}]

\begin{abstract}

Prior Arbitrary-Scale Image Super-Resolution (ASISR) methods often experience a significant performance decline when the upsampling factor exceeds the range covered by the training data, introducing substantial blurring. To address this issue, we propose a unified model, Stroke-based Cyclic Amplifier (SbCA), for ultra-large upsampling tasks.
The key of SbCA is the stroke vector amplifier, which decomposes the image into a series of strokes represented as vector graphics for magnification. Then, the detail completion module also restores missing details, ensuring high-fidelity image reconstruction. Our cyclic strategy achieves ultra-large upsampling by iteratively refining details with this unified SbCA model, trained only once for all, while keeping sub-scales within the training range.
Our approach effectively addresses the distribution drift issue and eliminates artifacts, noise and blurring, producing high-quality, high-resolution super-resolved images. Experimental validations on both synthetic and real-world datasets demonstrate that our approach significantly outperforms existing methods in ultra-large upsampling tasks  (e.g. $\times100$), delivering visual quality far superior to state-of-the-art techniques.

\end{abstract}

\section{Introduction}
\label{sec:intro}

Arbitrary-Scale Image Super-Resolution (ASISR) aims to upscale low-resolution (LR) images to high-resolution (HR) images at any arbitrary factor using a single model. While state-of-the-art ASISR methods \cite{liif, lte, linf, ciaosr, tsao2024boosting} achieve continuous and stable upsampling within the scaling range covered by the training data, their performance degrades significantly when the upsampling factor exceeds predefined range (the out-of-scale issue) , particularly when it extends far beyond the training distribution. This degradation results in suboptimal outcomes, severely limiting ASISR in ultra-large upsampling tasks, as illustrated in the lower part of Fig. \ref{fig:large}.
In theory, training on datasets with larger upsampling factors could alleviate this issue. However, as the scaling factor increases, mapping between LR and HR images becomes increasingly complex and ill-posed, leading to instability during model training \cite{glean}.

Cascade methods are commonly employed to handle fixed ultra-large upsampling tasks \cite{sr3, cascaded}. They decompose ultra-large upsampling tasks into smaller steps, training separate models for each, which are applied sequentially to achieve the final result. While cascade methods address the ill-posed nature of training a single ultra-large upsampling model, they have key drawbacks.
First, training is highly complex, as each step requires a dedicated model \cite{sr3} and tailored noise and blur strategies to mitigate distribution shifts \cite{cascaded}. Second, storage demands are high due to the need for multiple models. Third, cascade methods lack scalability, as new upsampling factors require additional models or retraining. These limitations make them impractical for arbitrary ultra-large upsampling tasks.

\begin{figure}[tp]
\centering
\begin{minipage}[h]{0.9\linewidth}
    \centering
    \begin{minipage}[h]{1\linewidth}
        \centering
        \begin{minipage}[h]{0.24\linewidth}
            \centering
            \scriptsize{LR}\\
            \vspace{0.05\linewidth}
            \includegraphics[width=1\linewidth]{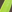}\\
        \end{minipage}
        \hfill
        \begin{minipage}[h]{0.24\linewidth}
            \centering
            \scriptsize{$\times$2}\\
            \vspace{0.05\linewidth}
            \includegraphics[width=1\linewidth]{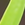}\\
        \end{minipage}
        \hfill
        \begin{minipage}[h]{0.24\linewidth}
            \centering
            \scriptsize{$\times$2$\times$2}\\
            \vspace{0.05\linewidth}
            \includegraphics[width=1\linewidth]{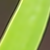}\\
        \end{minipage}
        \hfill
        \begin{minipage}[h]{0.24\linewidth}
            \centering
            \scriptsize{$\times$2$\times$2$\times$2}\\
           \vspace{0.05\linewidth}
            \includegraphics[width=1\linewidth]{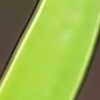}\\
        \end{minipage}\\
        \vspace{0.01\linewidth}
        \scriptsize{(a)  Artifact}
    \end{minipage}
    \vspace{0.001\linewidth}\\
    \begin{minipage}[h]{1\linewidth}
        \centering
        \begin{minipage}[h]{0.24\linewidth}
            \centering
            \includegraphics[width=1\linewidth]{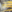}\\
        \end{minipage}
        \hfill
        \begin{minipage}[h]{0.24\linewidth}
            \centering
            \includegraphics[width=1\linewidth]{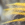}\\
        \end{minipage}
        \hfill
        \begin{minipage}[h]{0.24\linewidth}
            \centering
            \includegraphics[width=1\linewidth]{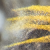}\\
        \end{minipage}
        \hfill
        \begin{minipage}[h]{0.24\linewidth}
            \centering
            \includegraphics[width=1\linewidth]{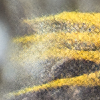}\\
        \end{minipage}\\
        \vspace{0.01\linewidth}
        \scriptsize{(b) Noise}
    \end{minipage}
    \vspace{0.001\linewidth}\\
    \begin{minipage}[h]{1\linewidth}
        \centering
        \begin{minipage}[h]{0.24\linewidth}
            \centering
            \includegraphics[width=1\linewidth]{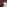}\\
        \end{minipage}
        \hfill
        \begin{minipage}[h]{0.24\linewidth}
            \centering
            \includegraphics[width=1\linewidth]{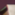}\\
        \end{minipage}
        \hfill
        \begin{minipage}[h]{0.24\linewidth}
            \centering
            \includegraphics[width=1\linewidth]{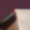}\\
        \end{minipage}
        \hfill
        \begin{minipage}[h]{0.24\linewidth}
            \centering
            \includegraphics[width=1\linewidth]{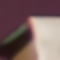}\\
        \end{minipage}\\
        \vspace{0.01\linewidth}
        \scriptsize{(c) Blur}
    \end{minipage}
\end{minipage}
\vspace{-0.1in}
\centering
\caption{Degradation is observed in existing ASISR methods when applied cyclically to uniformly scaled images.
\label{fig:ablation_2}}
\end{figure}

Building on the advantages of cascade models in addressing ill-posed problems through task decomposition, we aim to develop a unified ASISR model that can handle both small upsampling factors (e.g., $\times1 \sim 4$) and achieve ultra-large super-resolution through cyclic strategy. For example, $\times30$ can be progressively sampled by decomposing the task into smaller steps (e.g., $\times4$, $\times3$, $\times2.5$). The model can be applied iteratively to refine the resolution gradually.
However, severe distribution shift may occur when employing existing ASISR models \cite{liif, lte, linf, ciaosr, tsao2024boosting} cyclically, where the input progressively deviates from the original training distribution with each cycle. This can result in obvious artifacts, noise, and blurry boundaries (see Fig. \ref{fig:ablation_2}).

To address this issue, we propose a novel ASISR model for cyclic super-resolution approach, called Stroke-based Cyclic Amplifier (SbCA), to synthesize high-quality, high-resolution images. SbCA consists of two key components: the stroke vector amplifier and the detail completion module.
During each cycle, the input image first passes through the stroke vector amplifier, which decomposes the image into a series of strokes represented as vector graphics using Bézier curves. Based on the specified upsampling factor, the amplifier then magnifies and paints these strokes onto a canvas, generating an enlarged image. Theoretically, the complete set of strokes forms an overcomplete basis, enabling the reconstruction of any complex image content. The detail completion module subsequently refines the image by leveraging structural and texture priors, restoring fine details lost during the degradation process and ultimately enhancing perceptual quality. By cyclically applying our unified SbCA model, we achieve ultra-large upsampling factors, even up to $\times100$, as shown in Fig. \ref{fig:large}. 
Notably, SbCA is a unified model trained once on the $\times 1 \sim 4$ synthetic dataset, adapting to all upsampling factors and eliminating the need for separate training at each step or multiple rounds of training for the entire super-resolution process.

Furthermore, SbCA effectively mitigates the gradual accumulation of noise and artifacts and the progressive increase in edge blurring during cyclic processing. Since SbCA utilizes only a limited amount of vector-based information, noise and artifacts are inherently difficult to be learned and reproduced within such a sparse representation, thereby preventing the accumulation of artifacts. This is analogous to sparse coding reconstruction in image denoising \cite{le2005sparse, elad2006image}.
Additionally, SbCA inherits the desirable properties of vector graphics, which can be scaled to any size without quality loss, effectively suppressing edge blurring and ensuring high-quality upsampling results.

Our SbCA maintains stability throughout the cyclic process, preventing the accumulation and amplification of artifacts, thereby successfully overcoming the distribution drift problem. Our approach not only effectively resolves the ``out-of-scale" issue but also greatly reduces training and storage demands, as prior cascade methods require separate training for each upsampling step, whereas ours is trained once for all. Therefore, it enhances scalability for arbitrary upsampling factors.

The main contributions of our work are as follows:
\begin{itemize}
    \item We present a unified super-resolution model that applies cyclically to produce high-quality images at arbitrary ultra-large magnification scales, significantly outperforming SOTA methods in terms of visual quality.
    \item Our model effectively mitigates distribution drift and, despite being trained on synthetic datasets, generalizes well to real-world super-resolution tasks at ultra-large scales.
    \item Our approach reduces training and storage demands while improving scalability for arbitrary upsampling factors, making it practical and adaptable for diverse image super-resolution applications.

\end{itemize}

\section{Related Work}
\label{sec:relatedwork}


\subsection{Arbitrary-Scale Super-Resolution}

MetaSR \cite{metasr} introduced an innovative meta-upsampling module that dynamically generates filter weights for various upampling factors based on input coordinates and scales; however, its generalization ability for large-scale SR is limited. In contrast, LIIF \cite{liif} employs implicit neural representations, directly predicting the RGB values of corresponding SR image coordinates from local features of LR images using MLPs. By varying sampling density, it achieves SR results that surpass the training scale. LTE \cite{lte} further enhances implicit feature representation by introducing a local texture estimator, utilizing transformations in the Fourier domain. Additionally, some methods achieve SOTA performance by integrating attention mechanisms \cite{ciaosr, clit} or incorporating generative models \cite{linf,idm,tsao2024boosting}. Despite the impressive performance of implicit neural representation methods for ASISR capable of scaling up to $\times 30$—these models still encounter issues with blur or distortion when recovering fine image details in scenarios exceeding $\times 4$.

\subsection{Large-scale Image Super-Resolution}

\begin{figure*}[tp]
\centering
\includegraphics[scale=0.14]{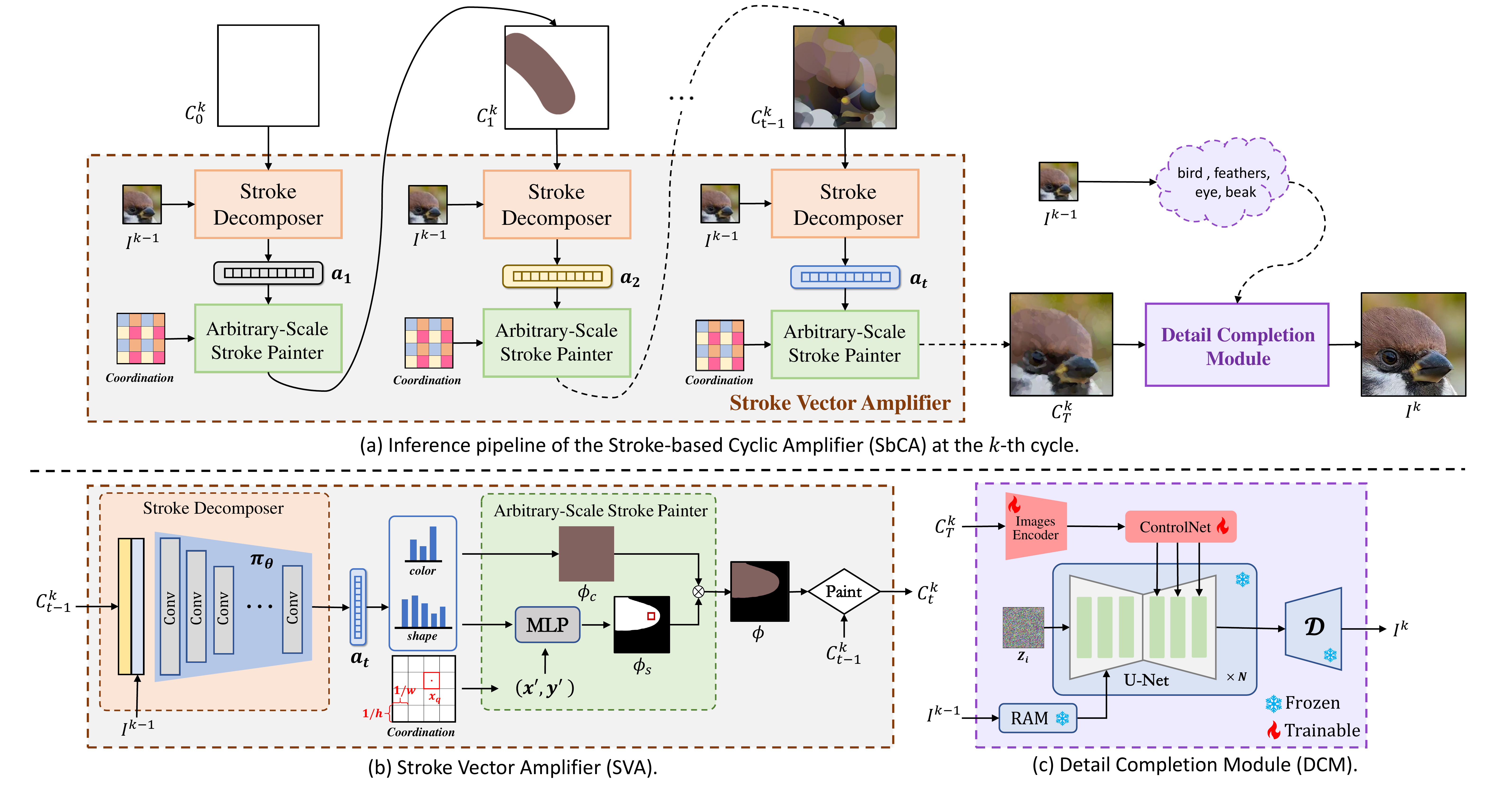} \\
\centering
\caption{
Illustration of our stroke-based cyclic amplifier framework: (a) Overview of the inference pipeline, (b) Detailed description of the components within the stroke vector amplifier, and (c) Internal structure of the detail completion module. Detailed process visualization can be found in the supplementary video.
}\label{fig:method}
\end{figure*}


Large-scale image SR has mainly been studied in the context of fixed SR factors. Many studies address the ill-posed nature of extreme downsampling by extending training datasets to contain LR-HR pairs up to scales of $\times 16$ or even $\times 64$, often restricting the task to specific semantic categories. For instance, the PULSE \cite{pulse} method confines the task to facial datasets, iteratively optimizing StyleGAN’s \cite{stylegan} latent encoding with pixel-wise constraints between the LR and SR outputs to generate HR faces similar to the input. GLEAN \cite{glean} extends this approach by conditioning a pre-trained StyleGAN generator on both latent codes and multi-resolution convolutional features, providing spatial guidance during SR and achieving scalability up to $\times 64$ across more diverse categories. Although these methods enhance fidelity at large scales in limited scenarios, they face challenges in generalizing to a wider range of real-world applications.

On the other hand, cascaded SR methods have recently gained traction for large-scale SR tasks without restricting the semantic content \cite{imagen,dell_e}. These methods achieve high upsampling factors by sequentially applying multiple SR models. For example, SR3 \cite{sr3} trains multiple $\times 4$ SR models in parallel and cascades them to achieve $\times 64$ SR. However, discrepancies between the distribution of the output from each preceding model and the training data hinder performance during cascading. CDM \cite{cascaded} addresses this by introducing noise and blur as data augmentations during each model’s training to reduce distribution gaps, effectively improving cascading performance. Yet, this approach incurs additional training costs and requires storage of multiple model parameters. Furthermore, meeting different upampling factors necessitates training multiple models, making scalability challenging. In summary, these methods are not suitable for arbitrary large-scale SR.

\section{Method}
\subsection{Overview}

Given an LR image and its expected upsampling factor \( s \), our method first decomposes \( s \) into a sequence of small upsampling factors, ensuring none exceed \( s_{\text{max}} \), the maximum upsampling factor used during training. Specifically, $s$ is decomposed as \( s = s^1 \times \cdots \times s^k \times \cdots \times s^K \), where \( s^k \leq s_{\text{max}} \), and \( k \) represents the order of these factors in the sequence, ranging from 1 to \( K \). Following this, the SbCA performs \( K \) consecutive upsampling operations according to this sequence, with the output of each operation serving as the input for the next. Specifically, during the \(k\)-th cycle, the amplifier takes the output image \( I^{k-1} \) from the previous step and scales it up to the factor specified by \( s^{k} \), generating a new intermediate image \( I^{k} \). The entire process starts with the original LR image \( I^0\) and goes through multiple cycles until the last operation is completed, resulting in the final HR output image \( I^K \).  

The key components of the SbCA include the Stroke Vector Amplifier and the Detail Completion Module. In the \( k \)-th cycle, as shown in Fig. \ref{fig:method}(a), the stroke vector amplifier the input image \( I^{k-1} \) into a stroke vector image \( C^k_T \) at the target resolution. The Detail Completion Module then enhances the detail features of \( C^k_T \), resulting in the final high-perceptual-quality output image \( I^k \). Next, we will provide a detailed description of the two core modules.

\subsection{Stroke Vector Amplifier (SVA)}
Given an image \( I^{k-1} \), the SVA decomposes it into total number of $T$ ordered stroke parameters. Next, decomposed stroke parameter is used to paint the stroke onto a blank canvas \( C^k_0 \) at the target resolution, producing the stroke image \( C^k_T \). The module consists of two components: the Stroke Decomposer, which parameterizes the strokes from the input image, and the Arbitrary-Scale Stroke Painter, which paints them into an image at arbitrary resolution.

\paragraph{Stroke Decomposer (SD).} 
This module decomposes the image into a series of strokes, formulated as a sequential decision-making problem and solved using reinforcement learning \cite{learning2paint}. Each stroke is parameterized by Bézier curves, with an agent responsible for the decomposition. As shown in Fig. \ref{fig:method}(b), given the input image \( I^{k-1} \), the canvas \( C^k_{t-1} \) after \( t-1 \) strokes, and the current step \( t \), the agent determines the current stroke’s parameters \( a_t = \{ x_0, y_0, x_1, y_1, x_2, y_2, w_0, w_1, r, g, b \} \) . Here, \( \{ x_0, y_0, x_1, y_1, x_2, y_2 \} \) are the Bézier control points, \( \{ w_0, w_1 \} \) are the stroke widths at the start and end, and \( \{ r, g, b \} \) represent the stroke color.

The agent uses a policy \( \pi_\theta \), where \( \theta \) represents the learnable parameters of a neural network, to map the current state \( \mathcal{S}_t \) to stroke parameters  \( a_t = \pi_\theta(\mathcal{S}_t) \). The state \( \mathcal{S}_t \) consists of three components: the input image \( I^{k-1} \), the  canvas \( C^k_{t-1} \), and the step number \( t \), with the canvas \( C^k_{t-1} \) downsampled to match the resolution of \( I^{k-1} \).

The goal of reinforcement learning is to adjust the learnable parameters \( \theta \) so that the agent's policy maximizes the cumulative reward:

\begin{equation}
    \pi_\theta^* = \arg\max_{\pi_\theta} \mathbb{E}_{\tau \sim \pi_\theta} \left( \sum^{T}_{t=0} \mathcal{R}_t(\tau) \right),
\end{equation}
where \( \tau \) represents the sequence of strokes generated, with a total length \( T \), which is proportional to the area of $I^{k-1}$(see Section \ref{sec:Implementation Details} for details). The reward function \( \mathcal{R}_t(\tau) \) is the reward function at the \( t \)-th step in the sequence. The reward is defined as the difference in scores between the next-step canvas \( C^k_{t} \) and the current canvas \( C^k_{t-1} \):

\begin{equation}
    \mathcal{R}_t = \mathcal{D}(C^k_{t}, I^{k}) - \mathcal{D}(C^k_{t-1}, I^{k}),
\end{equation}
where \( \mathcal{D}(\cdot) \) is the WGAN-GP \cite{wgan_gp} discriminator, used to assess the consistency of image content.

In this paper, we adopt an Actor-Critic framework \cite{ppo} to train the stroke decomposition agent.

\paragraph{Arbitrary-Scale Stroke Painter (ASSP).} 

Once the SD generates the stroke parameters \( a_t \), the ASSP needs to scale the stroke based on \( s^k \) and paint it onto the canvas, where \( s^k \) can be any floating-point value. Using neural networks to implicitly handle stroke scaling and painting is a viable approach. However, existing methods \cite{strokenet, stylized, learning2paint,hu2023stroke} for painting strokes into pixel space typically employ MLPs to map the parameters to a high-dimensional space, followed by stacked transpose convolutions for upsampling, mimicking the behavior of graphic engines. This architecture has a key limitation: once trained, the network's output dimensions are fixed due to the MLP, preventing the generation of strokes at arbitrary scales.

To address this limitation, we are inspired by implicit querying \cite{liif} that enables continuous SR of images. We propose the use of an implicit neural representation to model continuous stroke contours. By querying this representation at varying densities, we can achieve arbitrary-scale painting of stroke parameters. Our method queries pixel values at coordinates \( x_q = (x', y') \) on the target resolution canvas based on the stroke shape parameters \( a_{shape} \). By querying all coordinates of the target canvas, we obtain clear stroke  silhouettes representing the stroke parameters at the target resolution, as depicted in Fig. \ref{fig:method}(b). Given the shape parameters \( a_{shape} \), the target canvas resolution \( [h, w] \), and the set of query coordinates \( X_q \), the implicit querying process can be expressed as:
\begin{equation}
    \phi_s = r_\theta(a_{\text{shape}}, X_q, [c_h, c_w]),
\end{equation}
where \( r_\theta \) is an MLP-based stroke silhouette querying function, and \( [c_h, c_w] = [1/h, 1/w] \) denotes the size of each canvas pixel. The stroke silhouette \( \phi_s \) is thus obtained. Additionally, the stroke color is painted separately, with \( a_{color} \) expanded into a color tensor \( \phi_c \) of size \( [h, w, 3] \), which is then combined with the stroke silhouette \( \phi_s \) to synthesize the final rasterized stroke \( \phi \).




we employ an \( \mathcal{L}_1 \) loss to supervise the generation of stroke silhouettes \( \phi_s \):
\begin{equation}
    \mathcal{L}_1 = |\phi_s - \phi_{gt}|,
\end{equation}
where \( \phi_{gt} \) represents the ground-truth stroke silhouette, encouraging the ASSP to closely approximate it.

More details on the network architectures and model parameter sizes can be found in the appendix.

\subsection{Detail Completion Module (DCM)}
Although the SVA generates an upsampled painted image \( C^k_T \) with a scaling factor of \( s^k \), the painting process alone cannot restore the fine details lost during the degradation of the LR image. To enrich the details in the painted image, the DCM introduce a pre-trained generative model, such as the Diffusion Model (DM) \cite{ddpm}, to leverage its learned prior knowledge from natural images, adding fine details to \( C^k_T \) and thus generating high-perceptual-quality SR results.

However, during the cyclic SR process, using a pre-trained generative model presents two main challenges. First, it is necessary to ensure the consistency of the input image distribution for the generative model. Second, due to memory constraints, when performing ultra large-scale upsampling, the size of the image requiring detail completion increases with each cycle. As a result, we typically need to split the image into multiple patches, process each patch separately using the generative model for detail completion, and then stitch the completed patches together to form the final result. However, these image patches often lack rich contextual information, making it difficult for the generative model’s prior knowledge to be fully utilized.

The first challenge is addressed by the more stable image distribution generated by the SVA. The output image can be seen as a set of strokes selected from an infinite stroke space, where each stroke is free from noise, artifacts, and blurring. As a result, the image distribution produced by the SVA is highly stable.  With this issue resolved, we focus on the second challenge.

Recent advancements in multimodal technology, especially in image-to-text (I2T) alignment, offer a promising solution to this challenge. By leveraging I2T captioning and models such as CLIP \cite{clip}, we can generate textual descriptions of images that capture their contextual information. These descriptions can then serve as additional context for the generative model, helping to fill in gaps in the image patches and improve the completion of fine details.

In our approach, as illustrated in Fig. \ref{fig:method}(c), we use the I2T model RAM \cite{ram} to extract descriptive phrases \( p^{k-1}_{\text{txt}} \) from the previous image \( I^{k-1} \), thereby enhancing the contextual information of the image patches. Additionally, we employ ControlNet \cite{contorlnet} to encode the painted image \( C^{k}_T \). Both the extracted text descriptions and the encoded image are then fed as conditioning inputs into Stable Diffusion \cite{stablediffusion}, enabling the generation of high-perceptual-quality SR  results. The optimization process is formalized as follows:

\begin{equation}
    \mathcal{L} = \mathbb{E}_{z_0,i,p^{k-1}_{txt},C^{k}_T}[||\epsilon - \epsilon_\theta(z_i,i,p^{k-1}_{txt},C^{k}_T)||^2_2],
\end{equation}
where \( z_0 \) is the latent representation of \( I^{k} \), and \( z_i \) represents its noisy counterpart at a randomly selected diffusion step \( i \). Here, \( \epsilon_\theta \) predicts the noise added to \( z_i \).

It is worth noting that some existing SISR methods \cite{pasd, seesr} based on DM have incorporated semantic textual information. However, unlike these methods, which typically leverage both textual and image context simultaneously to help the network better utilize redundant information and improve SR performance, our approach relies solely on textual information to provide context. The goal is to help the network better understand the overall context of the image during detail recovery, leading to more accurate completion of finer details.
\begin{table*}[tbh]
    \scriptsize
    \setlength\tabcolsep{2pt}
    \centering
    \caption{Comparison with ASISR methods on the DIV8K synthetic dataset, with the best results in \textbf{bold}. Our approach significantly outperforms others in visual perceptual metrics (LPIPS, MUSIQ, NIQE, PI) while remaining competitive in PSNR and SSIM. \label{table:div8K-large}} 
    \resizebox{0.9\textwidth}{!}{
    \begin{tabular}{  l | c c c c c c| c c c c c c| c c c c c c}
        \toprule
        \multirow{3}{*}{Method}  &  \multicolumn{18}{c}{\textbf{DIV8K}}\\
        \multirow{2}{*}{} & \multicolumn{6}{c|}{$\times$8} & \multicolumn{6}{c|}{$\times$12} & \multicolumn{6}{c}{$\times$18}  \\

        \multirow{2}{*}{} & PSNR$\uparrow$  & SSIM$\uparrow$  & LPIPS$\downarrow$  & MUSIQ$\uparrow$ & NIQE$\downarrow$  &  PI$\downarrow$ & PSNR$\uparrow$ & SSIM$\uparrow$  &   LPIPS$\downarrow$ & MUSIQ$\uparrow$ & NIQE$\downarrow$  &  PI$\downarrow$ & PSNR$\uparrow$ & SSIM$\uparrow$  &  LPIPS$\downarrow$ & MUSIQ$\uparrow$ & NIQE$\downarrow$  &  PI$\downarrow$ \\

        \midrule
        LIIF \cite{liif}  & 28.75	&	0.760	&	0.457	&	20.03	&	9.74	&	8.62	&	26.92	&	0.725	&	0.542	&	19.26	&	11.11	&	9.36	&	25.63	&	0.706	&	0.592	&	19.54	&	12.32	&	10.08\\
        LTE \cite{lte}  & 28.83	&	0.760	&	0.460	&	19.97	&	9.84	&	8.69	&	27.02	&	0.725	&	0.547	&	19.60	&	11.34	&	9.48	&	25.65	&	0.706	&	0.596	&	19.66	&	12.63	&	10.32\\
        SRNO \cite{srno}  & 28.80	&	0.761	&	0.457	&	20.18	&	9.83	&	8.67	&	27.01	&	0.726	&	0.543	&	19.94	&	11.20	&	9.39	&	25.63	&	0.707	&	0.594	&	22.25	&	12.28	&	10.03\\
        EQSR \cite{eqsr} & 28.82	&	0.760	&	0.451	&	19.91	&	9.40	&	8.52	&	26.93	&	0.722	&	0.514	&	17.89	&	10.42	&	9.16	&	25.58	&	0.701	&	0.544	&	19.49	&	11.31	&	9.58\\
        CiaoSR \cite{ciaosr}  & \textbf{28.88}	&	\textbf{0.762}	&	0.455	&	20.02	&	9.78	&	8.65	&	\textbf{27.07}	&	\textbf{0.726}	&	0.540	&	19.62	&	11.17	&	9.40	&	25.68	&	\textbf{0.707}	&	0.591	&	20.22	&	12.37	&	10.12\\
         LINF \cite{linf}  & 28.83	&	0.760	&	0.442	&	19.81	&	9.22	&	8.10	&	27.06	&	0.724	&	0.528	&	19.59	&	10.50	&	8.67	&	25.64	&	0.705	&	0.578	&	20.30	&	12.49	&	9.51\\
         LINF-LP \cite{linf}  &	28.67	&	0.752	&	\textbf{0.399}	&	19.59	&	7.44	&	6.79 &	27.00	&	0.719	&	0.500	&	18.60	&	9.73	&	7.90&	\textbf{25.72}	&	0.702	&	0.561	&	18.99	&	13.28	&	9.70 \\
        SbCA (Ours)  & 25.02 & 0.662 & 0.424 & \textbf{37.26} & \textbf{4.26} & \textbf{4.45} & 24.02 & 0.650 & \textbf{0.459} & \textbf{38.29} & \textbf{4.25} & \textbf{4.48} & 23.09 & 0.645 & \textbf{0.494} & \textbf{35.72} & \textbf{4.76} & \textbf{4.75}\\	

        \midrule
        \multirow{2}{*}{Method} & \multicolumn{6}{c|}{$\times$24} & \multicolumn{6}{c|}{$\times$30} & \multicolumn{6}{c}{$\times 4 \times 3 \times 1.5$}  \\
        \multirow{2}{*}{} & PSNR$\uparrow$  & SSIM$\uparrow$  & LPIPS$\downarrow$  & MUSIQ$\uparrow$ & NIQE$\downarrow$  &  PI$\downarrow$ & PSNR$\uparrow$ & SSIM$\uparrow$  &   LPIPS$\downarrow$ & MUSIQ$\uparrow$ & NIQE$\downarrow$  &  PI$\downarrow$ & PSNR$\uparrow$ & SSIM$\uparrow$  &   LPIPS$\downarrow$ & MUSIQ$\uparrow$ & NIQE$\downarrow$  &  PI$\downarrow$ \\

        \midrule
        
        LIIF \cite{liif}  & 24.62	&	0.697	&	0.619	&	19.09	&	13.74	&	10.92	&	23.87	&	0.691	&	0.633	&	18.57	&	14.80	&	11.52          & 25.55 &	0.706 &	0.587 &	21.10 &	12.46 &	10.08	\\
        LTE \cite{lte}  & 24.68	&	0.697	&	0.622	&	19.01	&	14.11	&	11.19	&	23.94	&	\textbf{0.692}	&	0.635	&	18.55	&	15.05	&	11.67              & \textbf{25.59} &	\textbf{0.706} &	0.590 &	21.33 & 12.40 & 10.09	\\
        SRNO \cite{srno}  &24.62	&	0.697	&	0.620	&	20.63	&	13.53	&	10.75	&	23.85	&	0.691	&	0.634	&	19.30	&	14.36	&	11.15          & 25.34 &	0.705 &	0.586 &	22.12 &	12.46 &	10.15 \\
        EQSR \cite{eqsr} & 24.59	&	0.691	&	0.571	&	21.46	&	11.71	&	9.81	&	23.84	&	0.686	&	0.599	&	22.30	&	12.37	&	10.08          & 25.29 &	0.705 &	0.589 &	19.43 &	13.07 &	10.45	\\
        CiaoSR \cite{ciaosr}  & 24.68	&	\textbf{0.697}	&	0.618	&	19.74	&	13.78	&	10.96	&	23.93	&	0.692	&	0.633	&	18.91	&	14.87	&	11.54	   & 25.48 &	0.705 & 0.586 &	22.05 &	12.37 &	10.10 \\
         LINF \cite{linf}  & 24.66	&	0.695	&	0.608	&	19.87	&	14.42	&	10.12	&	23.90	&	0.690	&	0.625	&	19.53	&	15.25	&	10.30	       & 24.48 &	0.608 &	0.640 &	19.57 &	8.68 & 6.54\\
         LINF-LP \cite{linf}  &	\textbf{24.78}	&	0.695	&	0.594	&	19.41	&	15.46	&	10.86&	\textbf{24.04}	&	0.690	&	0.611	&	19.66	&	16.16	&	11.29&	14.96	&	0.534	&	0.772	&	20.19	&	10.32	&	7.66 \\
        SbCA (Ours)  & 22.30 & 0.633 & \textbf{0.501} & \textbf{38.10} & \textbf{4.46} & \textbf{4.51} & 21.62 & 0.622 & \textbf{0.522} & \textbf{38.56} & \textbf{4.49} & \textbf{4.53} & 23.09 & 0.645 & \textbf{0.494} & \textbf{35.72} & \textbf{4.76} & \textbf{4.75}\\	
        
        \bottomrule
    \end{tabular}
    }
\end{table*}

\section{Experiments}
\subsection{Dataset}
Following \cite{hat2023,rcan2018,2017ntire,eqsr}, we use the DF2K dataset \cite{2017ntire_datasets} as a training set and synthetically generate corresponding LR images via bicubic downsampling.

We tested our model on two types of datasets. First, to assess the performance of our proposed method for large-scale SR, we used the last 100 HR images from the DIV8K dataset \cite{div8k}. These HR images are downsampled with bicubic to synthesize the LR inputs. 

Additionally, to evaluate our method’s generalization capability to real-world images, we conducted experiments on three real-world datasets that progressively deviate from the bicubic training distribution: (1) The Benchmark, containing original high-quality images from (Set5 \cite{Set5}, Set14 \cite{Set14}, BSD100 \cite{BSD100}, and Urban100 \cite{urban100}), which are clear and unaltered by artificial degradation. (2) The RealSRSet \cite{realsrset} dataset, which is collected from the web and includes various forms of complex degradation, such as blur, compression artifacts, and noise. (3) The RealSR \cite{RealSR}, which consists of heavily degraded images captured by two cameras.


\subsection{Metrics}

We evaluate the perceptual performance of our model using LPIPS \cite{lpips} and no-reference image quality assessment metrics, including MUSIQ \cite{musiq}, NIQE \cite{NIQE}, and PI \cite{pi}. For real-world images, which typically lack ground truth references, we rely solely on no-reference metrics for evaluation. Additionally, we also use PSNR and SSIM as pixel-level metrics for reference.

\subsection{Implementation Details} \label{sec:Implementation Details}
In our experiments, \( s_{\text{max}} \) is set to 4, and the training process consists of three stages. First, we train the ASSP with a batch size of 64, where ground truth strokes are randomly generated by a graphics engine, with stroke sizes ranging from \( 16 \times 16 \) to \( 16s_{\text{max}} \times 16s_{\text{max}} \). Next, we fix the ASSP parameters and train the SVD module. The input LR image size is \( 16 \times 16 \), with a scaling factor from [1, \( s_{\text{max}} \)]. The training dataset is prepared using the paired data construction method from LIIF \cite{liif}. This stage has a batch size of 96, a replay buffer size of 16,000, a maximum of 20 inference steps, and an initial learning rate of \( 1 \times 10^{-4} \). Finally, using the trained SVA, we construct triplet data \(\{I_{\text{LR}}, C_T, I_{\text{HR}}\}\) to train the DCM. This stage uses a batch size of 32 and an initial learning rate of \( 5 \times 10^{-5} \). The entire training process took two days on A6000 GPU.


During the inference process, for an input image of size h $\times$ w, we first decompose it into \( 16 \times 16 \) patches and paint 20 strokes for each patch. Therefore, a total of \( \frac{h}{16} \times \frac{w}{16} \times 20 \) strokes are painted to reconstruct the entire image. After vector upsampling, the result is fed into the DCM to restore fine details before being processed in the next cycle.

\subsection{Comparisons with the State-of-the-Art}
We conducted comprehensive comparisons with several state-of-the-art arbitrary-scale super-resolution methods, including LIIF \cite{liif}, LTE \cite{lte}, SRNO \cite{srno}, CiaoSR \cite{ciaosr}, LINF \cite{linf} and LINF-LP \cite{tsao2024boosting}, across four diverse datasets. Since the publicly available checkpoint of the method in \cite{idm} does not support ASISR on natural images, we include a comparison with this method on the CelebA-HQ \cite{celeba_hq} face dataset in the appendix.



\begin{table*}[tbh]
    \scriptsize
    \setlength\tabcolsep{3.5pt}
    \centering
    \caption{Comparison with ASISR methods on real-world datasets, with the best results in \textbf{bold}. Our approach archives consistently superior performance over others, showcasing strong generalization in real-world image synthesis. \label{table:real-world-large}} 
    
    \resizebox{0.85\textwidth}{!}{
    \begin{tabular}{  l | c c c | c c c | c c c | c c c | c c c}      
        \toprule
        \multirow{3}{*}{Method}  &  \multicolumn{15}{c}{\textbf{Benchmark}}\\
        \multirow{2}{*}{} & \multicolumn{3}{c|}{$\times$8} & \multicolumn{3}{c|}{$\times$12} & \multicolumn{3}{c|}{$\times$18} & \multicolumn{3}{c|}{$\times$24} & \multicolumn{3}{c}{$\times$30}  \\

        \multirow{2}{*}{} & MUSIQ$\uparrow$ & NIQE$\downarrow$  &  PI$\downarrow$  & MUSIQ$\uparrow$  & NIQE$\downarrow$  &  PI$\downarrow$  & MUSIQ$\uparrow$   & NIQE$\downarrow$  &  PI$\downarrow$ & MUSIQ$\uparrow$ & NIQE$\downarrow$  &  PI$\downarrow$  & MUSIQ$\uparrow$ & NIQE$\downarrow$  &  PI$\downarrow$\\

        \midrule
        LIIF \cite{liif} &38.74	&	6.20	&	7.94	&	29.42	&	6.26	&	8.65	&	24.54	&	6.21	&	9.26	&	22.64	&	6.29	&	9.72	&	21.31	&	6.39	&	10.11\\
        LTE \cite{lte} & 37.62	&	6.32	&	8.06	&	27.81	&	6.47	&	8.83	&	23.58	&	6.39	&	9.54	&	22.41	&	6.48	&	10.04	&	21.64	&	6.57	&	10.47\\
        
        SRNO \cite{srno} & 38.55	&	6.20	&	8.00	&	29.21	&	6.07	&	8.74	&	24.40	&	5.65	&	9.28	&	22.94	&	5.52	&	9.57	&	21.39	&	5.65	&	9.86\\
        CiaoSR \cite{ciaosr} & 38.00	&	6.38	&	8.04	&	28.71	&	6.44	&	8.79	&	24.29	&	6.32	&	9.44	&	22.84	&	6.37	&	9.89	&	21.87	&	6.42	&	10.26\\
        LINF \cite{linf}& 37.25	&	6.40	&	8.13	&	27.57	&	6.23	&	8.81	&	23.36	&	5.84	&	9.40	&	22.29	&	5.84	&	9.80	&	21.32	&	5.94	&	10.13\\
        LINF-LP\cite{tsao2024boosting} &	30.65	&	6.15	&	5.55&	16.74	&	11.84	&	9.88&	18.54	&	14.80	&	11.26&	20.33	&	15.87	&	11.53&	20.99	&	16.10	&	11.45\\
        SbCA (Ours) & \textbf{68.81} & \textbf{4.23} & \textbf{4.46} & \textbf{64.01} & \textbf{4.06} & \textbf{4.56} & \textbf{55.60} & \textbf{4.10} & \textbf{4.78} & \textbf{46.08} & \textbf{4.01} & \textbf{4.96} & \textbf{42.06} & \textbf{4.06} & \textbf{4.93}\\

        \midrule
        
        \multirow{3}{*}{Method}  &  \multicolumn{15}{c}{\textbf{RealSRSet}}\\
        \multirow{2}{*}{} & \multicolumn{3}{c|}{$\times$8} & \multicolumn{3}{c|}{$\times$12} & \multicolumn{3}{c|}{$\times$18} & \multicolumn{3}{c|}{$\times$24} & \multicolumn{3}{c}{$\times$30}  \\

        \multirow{2}{*}{} & MUSIQ$\uparrow$ & NIQE$\downarrow$  &  PI$\downarrow$  & MUSIQ$\uparrow$  & NIQE$\downarrow$  &  PI$\downarrow$  & MUSIQ$\uparrow$   & NIQE$\downarrow$  &  PI$\downarrow$ & MUSIQ$\uparrow$ & NIQE$\downarrow$  &  PI$\downarrow$  & MUSIQ$\uparrow$ & NIQE$\downarrow$  &  PI$\downarrow$\\
        
        \midrule
        LIIF \cite{liif} & 31.01	&	6.25	&	7.75	&	26.16	&	6.16	&	8.37	&	24.17	&	6.06	&	8.70	&	22.56	&	6.08	&	9.29	&	21.59	&	6.17	&	9.67\\
        LTE \cite{lte} & 29.94	&	6.45	&	7.90	&	25.36	&	6.48	&	8.67	&	23.62	&	6.31	&	9.13	&	22.29	&	6.35	&	9.72	&	21.62	&	6.41	&	10.07\\
        
        SRNO \cite{srno} & 30.33	&	6.33	&	7.82	&	26.03	&	6.14	&	8.47	&	24.06	&	5.73	&	8.69	&	22.91	&	5.58	&	9.13	&	21.65	&	5.67	&	9.52\\
        CiaoSR \cite{ciaosr} & 32.90	&	5.95	&	7.87	&	25.51	&	5.99	&	8.55	&	24.50	&	5.93	&	8.88	&	22.93	&	5.98	&	9.42	&	22.36	&	6.15	&	9.74\\
        LINF \cite{linf}& 29.45	&	6.45	&	7.91	&	24.95	&	6.22	&	8.62	&	23.28	&	5.79	&	8.95	&	22.32	&	5.72	&	9.42	&	21.42	&	5.77	&	9.70\\
        LINF-LP\cite{tsao2024boosting} &	24.90	&	5.98	&	5.57&	19.49	&	7.80	&	6.51&	20.58	&	10.12	&	7.54&	20.78	&	12.81	&	8.90&	21.08	&	14.05	&	9.56\\
        SbCA (Ours) & \textbf{61.14} & \textbf{3.80} & \textbf{4.62} & \textbf{53.91} & \textbf{3.73} & \textbf{4.85} & \textbf{42.31} & \textbf{3.89} & \textbf{5.02} & \textbf{41.23} & \textbf{3.90} & \textbf{5.53} & \textbf{40.79} & \textbf{3.93} & \textbf{5.44}\\

        \midrule
        
        \multirow{3}{*}{Method}  &  \multicolumn{15}{c}{\textbf{RealSR}}\\
        \multirow{2}{*}{} & \multicolumn{3}{c|}{$\times$8} & \multicolumn{3}{c|}{$\times$12} & \multicolumn{3}{c|}{$\times$18} & \multicolumn{3}{c|}{$\times$24} & \multicolumn{3}{c}{$\times$30}\\

        \multirow{2}{*}{} & MUSIQ$\uparrow$ & NIQE$\downarrow$  &  PI$\downarrow$  & MUSIQ$\uparrow$  & NIQE$\downarrow$  &  PI$\downarrow$  & MUSIQ$\uparrow$   & NIQE$\downarrow$  &  PI$\downarrow$ & MUSIQ$\uparrow$ & NIQE$\downarrow$  &  PI$\downarrow$  & MUSIQ$\uparrow$ & NIQE$\downarrow$  &  PI$\downarrow$\\
        
        \midrule
        LIIF \cite{liif} &19.61	&	6.60	&	9.85	&	18.36	&	6.48	&	10.47	&	20.35	&	6.76	&	11.24	&	21.78	&	7.09	&	11.69	&	21.38	&	7.29	&	11.91\\
        LTE \cite{lte} & 19.51	&	6.68	&	9.97	&	18.33	&	6.57	&	10.60	&	20.34	&	6.85	&	11.40	&	21.81	&	7.18	&	11.84	&	21.42	&	7.38	&	12.04\\
        SRNO \cite{srno} & 19.60	&	6.67	&	9.94	&	18.34	&	6.46	&	10.54	&	20.35	&	6.70	&	11.31	&	21.84	&	6.85	&	11.69	&	21.46	&	7.15	&	11.86\\
        CiaoSR \cite{ciaosr} & 19.64	&	6.73	&	9.97	&	18.42	&	6.56	&	10.58	&	20.47	&	6.84	&	11.39	&	21.97	&	7.14	&	11.82	&	21.56	&	7.34	&	12.03\\
        LINF \cite{linf}& 19.58	&	6.66	&	9.96	&	18.40	&	6.53	&	10.58	&	20.42	&	6.78	&	11.38	&	21.87	&	7.08	&	11.80	&	21.47	&	7.28	&	11.97\\
        LINF-LP\cite{tsao2024boosting} &	18.27	&	9.91	&	8.96&	19.65	&	7.90	&	6.45&	18.34	&	10.89	&	8.00&	20.29	&	12.91	&	9.03&	21.01	&	13.85	&	9.57\\
        SbCA (Ours) & \textbf{59.28} & \textbf{4.77} & \textbf{5.73} & \textbf{55.45} & \textbf{4.68} & \textbf{5.71} & \textbf{47.51} & \textbf{4.95} & \textbf{6.07} & \textbf{41.88} & \textbf{4.86} & \textbf{6.13} & \textbf{39.27} & \textbf{4.92} & \textbf{5.94}\\
        \bottomrule

    \end{tabular}
    }
\end{table*}

\begin{figure*}[th]
\centering
\begin{minipage}[h]{0.85\linewidth}
    \begin{minipage}[h]{1\linewidth}
        \begin{minipage}[h]{0.2256\linewidth}
            \begin{minipage}[h]{1\linewidth}
                
                \begin{overpic}[width=1\linewidth, height=1.05\linewidth]{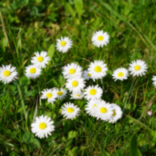}
                    \put(55,73){\includegraphics[width=0.13\linewidth]{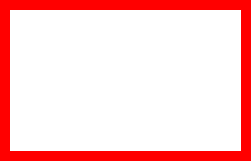}}
                \end{overpic}\\
                \centering
                \scriptsize{``DIV8K-1441" x30}
            \end{minipage}
        \end{minipage}
        \hfill
        \begin{minipage}[h]{0.18\linewidth}
            \begin{minipage}[h]{1\linewidth}
                \includegraphics[width=1\linewidth, height=0.6\linewidth]{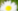} \\
                \centering
                \scriptsize{LR (PSNR$\uparrow$ / SSIM$\uparrow$)}
            \end{minipage}\\     
    
            \begin{minipage}[h]{1\linewidth}
                \includegraphics[width=1\linewidth, height=0.6\linewidth]{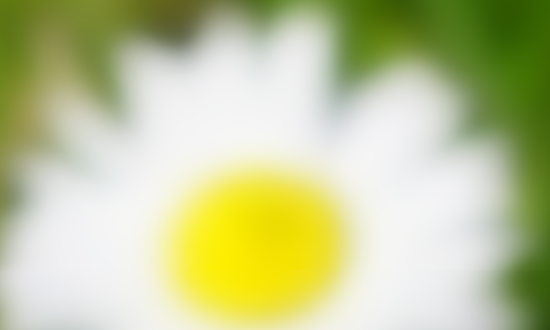} \\
                \centering
                \scriptsize{CiaoSR (23.80 / 0.917)}\\
            \end{minipage}\\
        \end{minipage}
        \hfill
        \begin{minipage}[h]{0.18\linewidth}
            \begin{minipage}[h]{1\linewidth}
                \includegraphics[width=1\linewidth, height=0.6\linewidth]{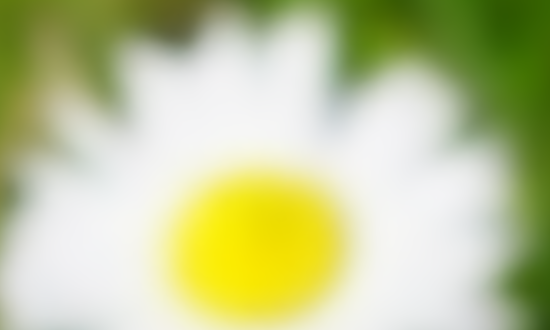} \\
                \centering
                \scriptsize{LIIF (23.85 / 0.916)}\\
            \end{minipage}\\     
    
            \begin{minipage}[h]{1\linewidth}
                \includegraphics[width=1\linewidth, height=0.6\linewidth]{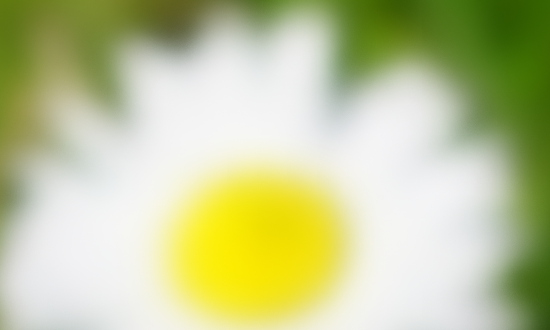} \\
                \centering
                \scriptsize{LINF (23.70 / 0.914)}
            \end{minipage}\\
        \end{minipage}
        \hfill
        \begin{minipage}[h]{0.18\linewidth}
            \begin{minipage}[h]{1\linewidth}
                \includegraphics[width=1\linewidth, height=0.6\linewidth]{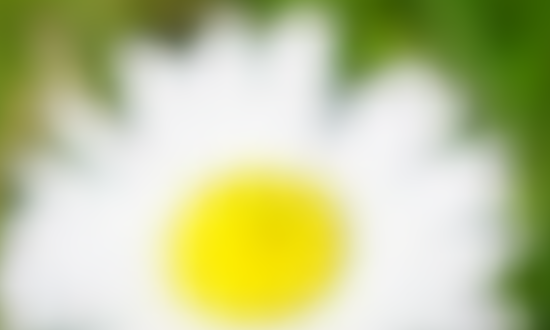} \\
                \centering
                \scriptsize{LTE (23.57 / 0.913)}\\
            \end{minipage}\\     
    
            \begin{minipage}[h]{1\linewidth}
                \includegraphics[width=1\linewidth, height=0.6\linewidth]{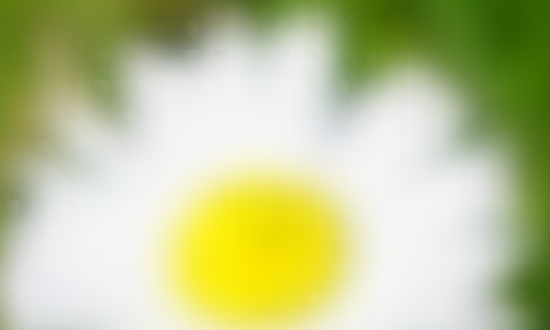} \\
                \centering
                \scriptsize{LINF-LP (22.82 / 0.903)}\\
            \end{minipage}\\
        \end{minipage}
        \hfill
        \begin{minipage}[h]{0.18\linewidth}
            \begin{minipage}[h]{1\linewidth}
                \includegraphics[width=1\linewidth, height=0.6\linewidth]{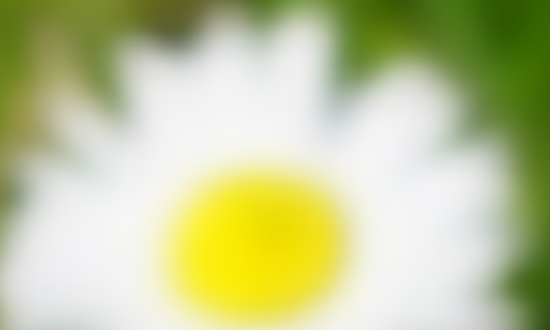} \\
                \centering
                \scriptsize{SRNO (23.76 / 0.916)}\\
            \end{minipage}\\     
    
            \begin{minipage}[h]{1\linewidth}
                \includegraphics[width=1\linewidth, height=0.6\linewidth]{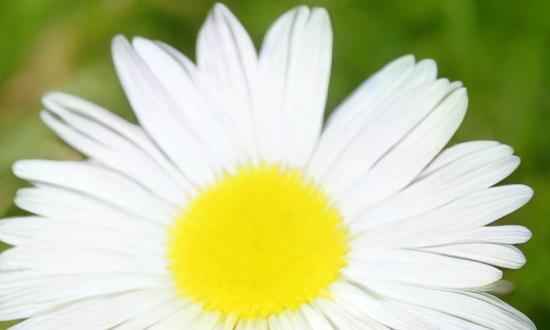} \\
                \centering
                \scriptsize{SbCA (15.29 / 0.585)}
            \end{minipage}\\
        \end{minipage}
    \end{minipage}

    \begin{minipage}[h]{1\linewidth}
        \begin{minipage}[h]{0.2256\linewidth}
            \begin{minipage}[h]{1\linewidth}
                \begin{overpic}[width=1\linewidth, height=1.05\linewidth]{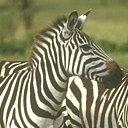}
                    \put(22,67){\includegraphics[width=0.18\linewidth]{resource/exp_arbsr/box.png}}
                \end{overpic}\\
                \centering
                \scriptsize{``BSD100-img\_052" x24}
            \end{minipage}
        \end{minipage}
        \hfill
        \begin{minipage}[h]{0.18\linewidth}
            \begin{minipage}[h]{1\linewidth}
                \includegraphics[width=1\linewidth, height=0.6\linewidth]{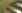} \\
                \centering
                \scriptsize{LR}
            \end{minipage}\\     
    
            \begin{minipage}[h]{1\linewidth}
                \includegraphics[width=1\linewidth, height=0.6\linewidth]{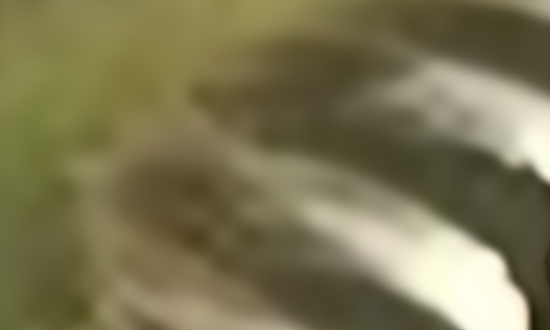} \\
                \centering
                \scriptsize{CiaoSR}
            \end{minipage}\\
        \end{minipage}
        \hfill
        \begin{minipage}[h]{0.18\linewidth}
            \begin{minipage}[h]{1\linewidth}
                \includegraphics[width=1\linewidth, height=0.6\linewidth]{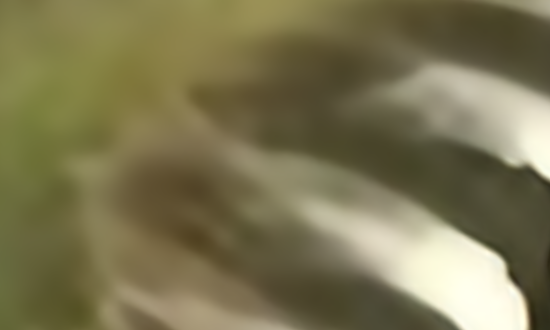} \\
                \centering
                \scriptsize{LIIF}\\
            \end{minipage}\\     
    
            \begin{minipage}[h]{1\linewidth}
                \includegraphics[width=1\linewidth, height=0.6\linewidth]{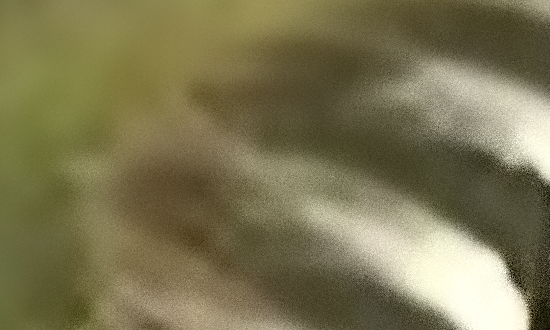} \\
                \centering
                \scriptsize{LINF}
            \end{minipage}\\
        \end{minipage}
        \hfill
        \begin{minipage}[h]{0.18\linewidth}
            \begin{minipage}[h]{1\linewidth}
                \includegraphics[width=1\linewidth, height=0.6\linewidth]{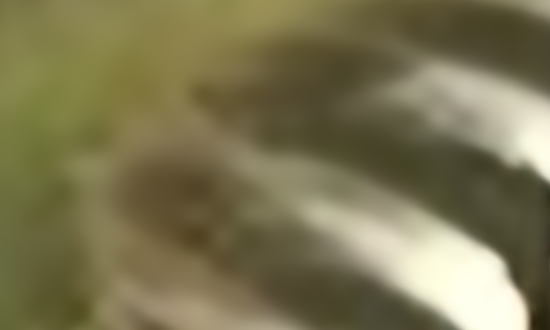} \\
                \centering
                \scriptsize{LTE}\\
            \end{minipage}\\     
    
            \begin{minipage}[h]{1\linewidth}
                \includegraphics[width=1\linewidth, height=0.6\linewidth]{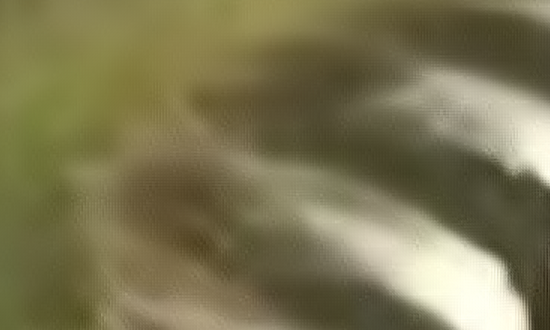} \\
                \centering
                \scriptsize{LINF-LP}
            \end{minipage}\\
        \end{minipage}
        \hfill
        \begin{minipage}[h]{0.18\linewidth}
            \begin{minipage}[h]{1\linewidth}
                \includegraphics[width=1\linewidth, height=0.6\linewidth]{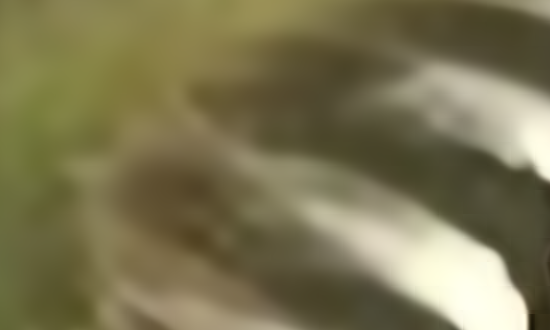} \\
                \centering
                \scriptsize{SRNO}\\
            \end{minipage}\\     
    
            \begin{minipage}[h]{1\linewidth}
                \includegraphics[width=1\linewidth, height=0.6\linewidth]{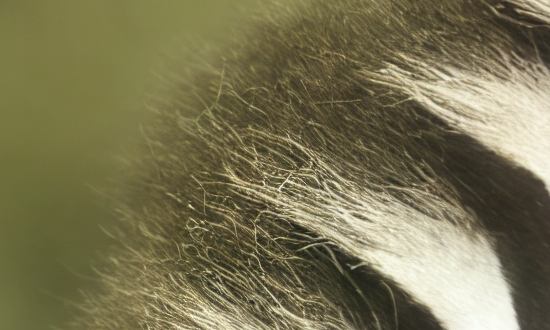} \\
                \centering
                \scriptsize{SbCA}
            \end{minipage}\\
        \end{minipage}
    \end{minipage}

\end{minipage}

\centering
\caption{Qualitative comparison with different methods. More qualitative results can be found in the supplementary video.  \label{fig:arb-scale}} 
\end{figure*}

\subsubsection{Experiments on Synthetic Dataset}
Table \ref{table:div8K-large} provides a detailed comparison of our proposed method against baseline approaches on the synthetic DIV8K dataset.  Specifically, our method achieves the best perceptual performance across all upsampling factors, particularly at the $\times 30$ upsampling factor, where it outperforms the second-best method LINF-LP by 14.6\% in terms of LPIPS. Additionally, for no-reference perceptual metrics such as MUSIQ, NIQE, and PI, our method also outperforms the second-best method LINF by 96.4\%, 70.5\%, and 55.8\%, respectively. As the upsampling factor increases, our method maintains stable performance across all perceptual metrics, while LINF-LP, despite its advantages at lower scales, deteriorates significantly at larger ones. These results highlight our approach's state-of-the-art perceptual performance in large-scale arbitrary super-resolution. 

It is worth noting that, we provide PSNR and SSIM as reference metrics, as they do not always align well with human visual perception \cite{realesrgan,wang2021real, wang2021unsupervised}. As shown in Fig. \ref{fig:arb-scale}, while our method lags in PSNR and SSIM, it achieves superior visual quality. To further validate this, we conducted a user study with 20 participants evaluating 100 images across four datasets. 86.88\% of participants preferred our method. We provide these details in the appendix.

\label{sec:4.4.1}Additionally, to demonstrate the issue of distribution drift during the  cyclic process in baseline methods, we take the \( \times 18 \) upsampling as an example and decompose it into a cyclic sequence of \( \times 4 \), \( \times 3 \), and \( \times 1.5 \), as shown in Table \ref{table:div8K-large}. Although each scaling step remains within the ``in-scale", the overall performance does not show significant improvement compared to direct \( \times 18 \) upsampling. This validates the impact of distribution shifts caused by the cyclic process on performance and further corroborates the effectiveness of our proposed method in mitigating these variations.

\subsubsection{Experiments on Real-World Datasets}
Table \ref{table:real-world-large} presents a comparison of all methods on real-world datasets. Due to the absence of ground-truth references, we utilized no-reference perceptual metrics for evaluation. The results show that our method consistently achieves superior perceptual quality. Specifically: (1) When dataset distributions deviate from the training distribution, our method outperforms baseline methods significantly. For example, on the benchmark dataset at a $\times 30$ upsampling factor, our method improves the three perceptual metrics by 96.6\%, 28.1\%, and 50\%, respectively, over the second-best method SRNO. (2) As the divergence between test dataset and training distribution increases, baseline performance declines. On the RealSR dataset, where distribution shift is most pronounced, baseline performance drops sharply, whereas our method remains largely unaffected. These findings confirm that our approach is both robust and effective under varying distribution conditions.

\subsubsection{Qualitative Evaluation}
Fig. \ref{fig:arb-scale} presents qualitative results comparing our method with baseline approaches across various datasets and upsampling factors. In the first row, for the extreme $\times 30$ upsampling factor, all baseline methods struggle to recover clear petal details, while the proposed method successfully preserves sharp and well-defined petal edges. In the second row, although LIIF and CiaoSR partially recover the zebra stripe edges, they fail to capture the fur details. Moreover, LINF and LINF-LP introduce additional artifacts and noise. In contrast, the proposed method generates realistic and sharp fur textures. 

\subsubsection{Timecost Evaluation}
We analyzed the inference time of various ASSR methods. Specifically, we set
the input size \(128 \times 128\) and performing $\times4$ upsampling, measuring the inference time on an NVIDIA A6000. The results are presented in Table \ref{table:infer}.  

Although our method requires an inference time of 2.22s, it achieves superior visual quality and is acceptable. Notably, a significant portion (1.86s) is attributed to the diffusion-based DCM, which involves a 20-step denoising process using DDIM \cite{ddim}. In future work, our aim is to improve efficiency by exploring one-step diffusion models.
\begin{table}[htpb]
    \scriptsize
    \setlength\tabcolsep{3.6pt}
    \centering
    \caption{Comparison of inference time across different models. \label{table:infer}} 

    \begin{tabular}{  c | c c c c c c c}
        
        \toprule
        \textbf{Models}   & \textbf{LIIF} & \textbf{LTE} & \textbf{SRNO} & \textbf{CiaoSR} & \textbf{LINF} & \textbf{LINF-LP}  & \textbf{SbCA}\\
        \midrule
        \textbf{Infer Time (s)} & 0.13 & 0.29 & 0.12 & 1.01 & 0.49 & 0.11 & 2.22\\

        \bottomrule
    \end{tabular}
\end{table}

\subsection{Ablation Study} 



To assess each component's impact in SbCA, we designed three variants: \textbf{DCM only}, \textbf{SVA only}, and the full model \textbf{SVA+DCM}, with ablation experiments following the decomposition method in Section \ref{sec:4.4.1}.

\begin{figure}[tp]
\centering
\begin{minipage}[h]{0.95\linewidth}
    \begin{minipage}[h]{1\linewidth}
        \centering
        \begin{minipage}[h]{0.19\linewidth}
            \centering
            \includegraphics[width=1\linewidth]{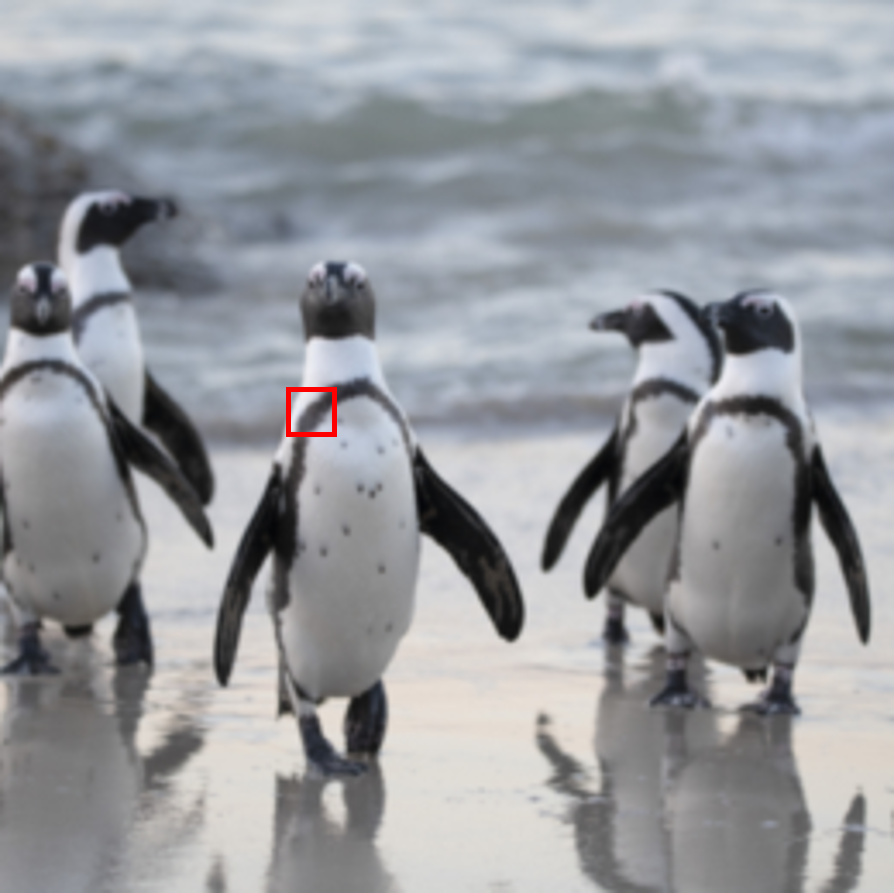}\\  
        \end{minipage}
        \hfill
        \begin{minipage}[h]{0.19\linewidth}
            \centering
            \includegraphics[width=1\linewidth]{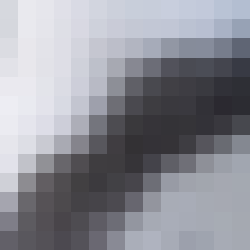}\\    
        \end{minipage}
        \hfill
        \begin{minipage}[h]{0.19\linewidth}
            \centering
            \includegraphics[width=1\linewidth]{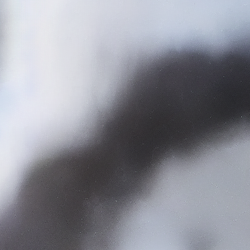}\\
        \end{minipage}
        \hfill
        \begin{minipage}[h]{0.19\linewidth}
            \centering
            \includegraphics[width=1\linewidth]{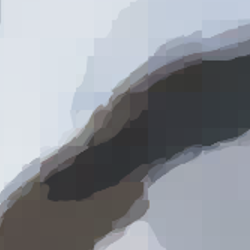}\\
        \end{minipage}
        \hfill
        \begin{minipage}[h]{0.19\linewidth}
            \centering
            \includegraphics[width=1\linewidth]{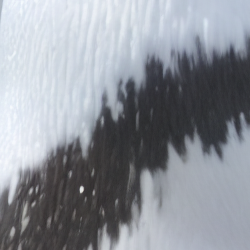}
        \end{minipage}
    \end{minipage}
    \vspace{0.001\linewidth}\\
    \begin{minipage}[h]{1\linewidth}
        \centering
        \begin{minipage}[h]{0.19\linewidth}
            \centering
            \includegraphics[width=1\linewidth]{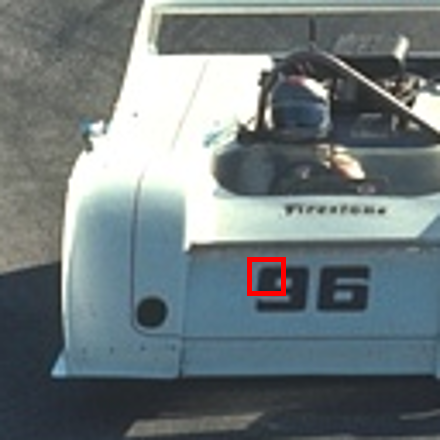}\\
            \scriptsize{Origin}
        \end{minipage}
        \hfill
        \begin{minipage}[h]{0.19\linewidth}
            \centering
            \includegraphics[width=1\linewidth]{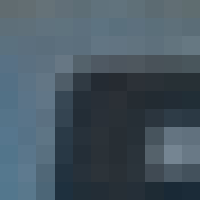}\\
            \scriptsize{LR}
        \end{minipage}
        \hfill
        \begin{minipage}[h]{0.19\linewidth}
            \centering
            \includegraphics[width=1\linewidth]{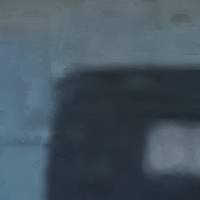}\\
            \scriptsize{\textbf{DCM only} }
        \end{minipage}
        \hfill
        \begin{minipage}[h]{0.19\linewidth}
            \centering
            \includegraphics[width=1\linewidth]{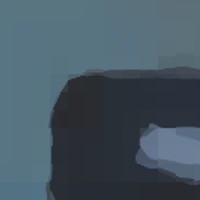}\\
            \scriptsize{\textbf{SVA only} }
        \end{minipage}
        \hfill
        \begin{minipage}[h]{0.19\linewidth}
            \centering
            \includegraphics[width=1\linewidth]{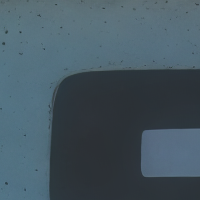}\\
            \scriptsize{\textbf{SVA}+\textbf{DCM}}
    \end{minipage}
    \end{minipage}
\end{minipage}

\centering
\caption{ 
Impact of different components. Using only the diffusion model (DCM) causes blurriness and artifacts due to distribution shifts, hindering fine details. Without DCM, outputs lack realism. Combining both ensures sharp, realistic results.}\label{fig:ablation_1}
\end{figure}

Table \ref{table:abla_1} shows the impact of each component on SR performance. Using only the diffusion model DCM in a cyclic manner results in suboptimal outcomes due to distribution shifts, causing blurry penguin feathers and digit artifacts (Fig. \ref{fig:ablation_1}). Using only the SVA module leads to texture loss and unrealistic outputs. In contrast, the full model restores fine feather textures and clear digit edges while eliminating artifacts, highlighting the SVA module's role in stabilizing distribution and the importance of detail completion for enhancing visual quality.  More ablation experiments can be found in the appendix.


\begin{table}[tp]
    \caption{Ablation study of various components. The best performance is achieved when all components are enabled.
    \label{table:abla_1}}
    \centering
    \scriptsize
    \resizebox{0.95\linewidth}{!}{
    \begin{tabular}{c | c  | c c c c } 
        \toprule
        \multirow{2}{*}{\textbf{Datasets}} & \multirow{2}{*}{\textbf{Model Variations}} & \multicolumn{4}{c}{\textbf{$\times$ 18} ($\times 4 \times 3 \times 1.5$)} \\ 
        {} & {}  & LPIPS$\downarrow$ & MUSIQ$\uparrow$  & NIQE$\downarrow$  & PI$\downarrow$  \\
        \midrule
        \multirow{3}{*}{DIV8K}&  \textbf{DCM only}  & 0.533 &	19.00 &	9.36 &	7.26	 \\ 
    
        {} & \textbf{SVA only} &0.569 &	27.13 &	16.15 &	11.25		\\  

        
        {}&\textbf{SVA + DCM} & \textbf{0.494} &	\textbf{35.72} &	\textbf{4.76} &	\textbf{4.75}	\\      	
        
        \midrule
        
        \multirow{3}{*}{BSD100} & \textbf{DCM  only} & - & 38.74	&	4.72	&	5.03	 \\ 
    
        {} &  \textbf{SVA  only} & - &	38.65 &	9.01 &	10.54		\\


        {}&\textbf{SVA + DCM}  & -	& \textbf{57.64}	&	\textbf{3.80}	&	\textbf{4.36}	\\     
        \bottomrule
    \end{tabular}
    }
\end{table}

\section{Conclusion}

In this paper, we presented a unified super-resolution model designed for scalable, high-quality ultra-large upsampling. Unlike traditional methods, our SbCA maintains stability through cyclic processing, effectively preventing the accumulation of artifacts, noise, and blurring, as well as distribution drift, ensuring superior visual quality. We highlight that SbCA is a unified model available for any upsampling factor that eliminates the need for separate training at different scales, significantly reducing computational and storage costs while enhancing scalability. Our approach not only outperforms existing methods in ultra-large super-resolution tasks but also generalizes well to real-world applications, making it a practical solution for diverse image upsampling applications. With its ability to handle arbitrary magnification factors, the SbCA model paves the way for future advancements in image restoration, enhancement, and beyond.

{
    \small
    \bibliographystyle{ieeenat_fullname}
    \bibliography{main}

\begin{thebibliography}{52}
\providecommand{\natexlab}[1]{#1}
\providecommand{\url}[1]{\texttt{#1}}
\expandafter\ifx\csname urlstyle\endcsname\relax
  \providecommand{\doi}[1]{doi: #1}\else
  \providecommand{\doi}{doi: \begingroup \urlstyle{rm}\Url}\fi

\bibitem[Agustsson and Timofte(2017)]{2017ntire_datasets}
Eirikur Agustsson and Radu Timofte.
\newblock Ntire 2017 challenge on single image super-resolution: Dataset and study.
\newblock In \emph{Proceedings of the IEEE conference on computer vision and pattern recognition workshops}, pages 126--135, 2017.

\bibitem[Bevilacqua et~al.(2012)Bevilacqua, Roumy, Guillemot, and Alberi-Morel]{Set5}
Marco Bevilacqua, Aline Roumy, Christine Guillemot, and Marie~Line Alberi-Morel.
\newblock Low-complexity single-image super-resolution based on nonnegative neighbor embedding.
\newblock In \emph{Proceedings of the British Machine Vision Conference (BMVC)}, 2012.

\bibitem[Blau et~al.(2018)Blau, Mechrez, Timofte, Michaeli, and Zelnik-Manor]{pi}
Yochai Blau, Roey Mechrez, Radu Timofte, Tomer Michaeli, and Lihi Zelnik-Manor.
\newblock The 2018 pirm challenge on perceptual image super-resolution.
\newblock In \emph{Proceedings of the European conference on computer vision (ECCV) workshops}, pages 0--0, 2018.

\bibitem[Cai et~al.(2019)Cai, Zeng, Yong, Cao, and Zhang]{RealSR}
Jianrui Cai, Hui Zeng, Hongwei Yong, Zisheng Cao, and Lei Zhang.
\newblock Toward real-world single image super-resolution: A new benchmark and a new model.
\newblock In \emph{In ICCV}, pages 3086--3095, 2019.

\bibitem[Cao et~al.(2023)Cao, Wang, Xian, Li, Ni, Pi, Zhang, Zhang, Timofte, and Van~Gool]{ciaosr}
Jiezhang Cao, Qin Wang, Yongqin Xian, Yawei Li, Bingbing Ni, Zhiming Pi, Kai Zhang, Yulun Zhang, Radu Timofte, and Luc Van~Gool.
\newblock Ciaosr: Continuous implicit attention-in-attention network for arbitrary-scale image super-resolution.
\newblock In \emph{In CVPR}, pages 1796--1807, 2023.

\bibitem[Chan et~al.(2021)Chan, Wang, Xu, Gu, and Loy]{glean}
Kelvin~CK Chan, Xintao Wang, Xiangyu Xu, Jinwei Gu, and Chen~Change Loy.
\newblock Glean: Generative latent bank for large-factor image super-resolution.
\newblock In \emph{In CVPR}, pages 14245--14254, 2021.

\bibitem[Chen et~al.(2022)Chen, Shi, Qin, Li, Han, Yang, and Guo]{realesrgan}
Chaofeng Chen, Xinyu Shi, Yipeng Qin, Xiaoming Li, Xiaoguang Han, Tao Yang, and Shihui Guo.
\newblock Real-world blind super-resolution via feature matching with implicit high-resolution priors.
\newblock In \emph{Proceedings of the 30th ACM International Conference on Multimedia}, pages 1329--1338, 2022.

\bibitem[Chen et~al.(2023{\natexlab{a}})Chen, Xu, Hong, Tsai, Kuo, and Lee]{clit}
Hao-Wei Chen, Yu-Syuan Xu, Min-Fong Hong, Yi-Min Tsai, Hsien-Kai Kuo, and Chun-Yi Lee.
\newblock Cascaded local implicit transformer for arbitrary-scale super-resolution.
\newblock In \emph{In CVPR}, pages 18257--18267, 2023{\natexlab{a}}.

\bibitem[Chen et~al.(2023{\natexlab{b}})Chen, Wang, Zhou, Qiao, and Dong]{hat2023}
Xiangyu Chen, Xintao Wang, Jiantao Zhou, Yu Qiao, and Chao Dong.
\newblock Activating more pixels in image super-resolution transformer.
\newblock In \emph{In CVPR}, pages 22367--22377, 2023{\natexlab{b}}.

\bibitem[Chen et~al.(2021)Chen, Liu, and Wang]{liif}
Yinbo Chen, Sifei Liu, and Xiaolong Wang.
\newblock Learning continuous image representation with local implicit image function.
\newblock In \emph{In CVPR}, pages 8628--8638, 2021.

\bibitem[Elad and Aharon(2006)]{elad2006image}
Michael Elad and Michal Aharon.
\newblock Image denoising via sparse and redundant representations over learned dictionaries.
\newblock \emph{IEEE Transactions on Image processing}, 15\penalty0 (12):\penalty0 3736--3745, 2006.

\bibitem[Gao et~al.(2023)Gao, Liu, Zeng, Xu, Li, Luo, Liu, Zhen, and Zhang]{idm}
Sicheng Gao, Xuhui Liu, Bohan Zeng, Sheng Xu, Yanjing Li, Xiaoyan Luo, Jianzhuang Liu, Xiantong Zhen, and Baochang Zhang.
\newblock Implicit diffusion models for continuous super-resolution.
\newblock In \emph{In CVPR}, pages 10021--10030, 2023.

\bibitem[Gu et~al.(2019)Gu, Lugmayr, Danelljan, Fritsche, Lamour, and Timofte]{div8k}
Shuhang Gu, Andreas Lugmayr, Martin Danelljan, Manuel Fritsche, Julien Lamour, and Radu Timofte.
\newblock Div8k: Diverse 8k resolution image dataset.
\newblock In \emph{2019 IEEE/CVF International Conference on Computer Vision Workshop (ICCVW)}, pages 3512--3516. IEEE, 2019.

\bibitem[Gulrajani et~al.(2017)Gulrajani, Ahmed, Arjovsky, Dumoulin, and Courville]{wgan_gp}
Ishaan Gulrajani, Faruk Ahmed, Martin Arjovsky, Vincent Dumoulin, and Aaron~C Courville.
\newblock Improved training of wasserstein gans.
\newblock \emph{Advances in neural information processing systems}, 30, 2017.

\bibitem[Ho et~al.(2020)Ho, Jain, and Abbeel]{ddpm}
Jonathan Ho, Ajay Jain, and Pieter Abbeel.
\newblock Denoising diffusion probabilistic models.
\newblock \emph{Advances in neural information processing systems}, 33:\penalty0 6840--6851, 2020.

\bibitem[Ho et~al.(2022)Ho, Saharia, Chan, Fleet, Norouzi, and Salimans]{cascaded}
Jonathan Ho, Chitwan Saharia, William Chan, David~J Fleet, Mohammad Norouzi, and Tim Salimans.
\newblock Cascaded diffusion models for high fidelity image generation.
\newblock \emph{Journal of Machine Learning Research}, 23\penalty0 (47):\penalty0 1--33, 2022.

\bibitem[Hu et~al.(2023)Hu, Yi, Zhu, Liu, Peng, Wang, Wang, and Ma]{hu2023stroke}
Teng Hu, Ran Yi, Haokun Zhu, Liang Liu, Jinlong Peng, Yabiao Wang, Chengjie Wang, and Lizhuang Ma.
\newblock Stroke-based neural painting and stylization with dynamically predicted painting region.
\newblock In \emph{Proceedings of the 31st ACM International Conference on Multimedia}, pages 7470--7480, 2023.

\bibitem[Hu et~al.(2019)Hu, Mu, Zhang, Wang, Tan, and Sun]{metasr}
Xuecai Hu, Haoyuan Mu, Xiangyu Zhang, Zilei Wang, Tieniu Tan, and Jian Sun.
\newblock Meta-sr: A magnification-arbitrary network for super-resolution.
\newblock In \emph{In CVPR}, pages 1575--1584, 2019.

\bibitem[Huang et~al.(2015)Huang, Singh, and Ahuja]{urban100}
Jia-Bin Huang, Abhishek Singh, and Narendra Ahuja.
\newblock Single image super-resolution from transformed self-exemplars.
\newblock In \emph{Proceedings of the IEEE conference on computer vision and pattern recognition}, pages 5197--5206, 2015.

\bibitem[Huang et~al.(2019)Huang, Heng, and Zhou]{learning2paint}
Zhewei Huang, Wen Heng, and Shuchang Zhou.
\newblock Learning to paint with model-based deep reinforcement learning.
\newblock In \emph{In ICCV}, pages 8709--8718, 2019.

\bibitem[Karras et~al.(2017)Karras, Aila, Laine, and Lehtinen]{celeba_hq}
Tero Karras, Timo Aila, Samuli Laine, and Jaakko Lehtinen.
\newblock Progressive growing of gans for improved quality, stability, and variation.
\newblock \emph{arXiv preprint arXiv:1710.10196}, 2017.

\bibitem[Karras et~al.(2019)Karras, Laine, and Aila]{stylegan}
Tero Karras, Samuli Laine, and Timo Aila.
\newblock A style-based generator architecture for generative adversarial networks.
\newblock In \emph{In CVPR}, pages 4401--4410, 2019.

\bibitem[Ke et~al.(2021)Ke, Wang, Wang, Milanfar, and Yang]{musiq}
Junjie Ke, Qifei Wang, Yilin Wang, Peyman Milanfar, and Feng Yang.
\newblock Musiq: Multi-scale image quality transformer.
\newblock In \emph{In ICCV}, pages 5148--5157, 2021.

\bibitem[Le~Pennec and Mallat(2005)]{le2005sparse}
Erwan Le~Pennec and St{\'e}phane Mallat.
\newblock Sparse geometric image representations with bandelets.
\newblock \emph{IEEE transactions on image processing}, 14\penalty0 (4):\penalty0 423--438, 2005.

\bibitem[Lee and Jin(2022)]{lte}
Jaewon Lee and Kyong~Hwan Jin.
\newblock Local texture estimator for implicit representation function.
\newblock In \emph{In CVPR}, pages 1929--1938, 2022.

\bibitem[Martin et~al.(2001)Martin, Fowlkes, Tal, and Malik]{BSD100}
David Martin, Charless Fowlkes, Doron Tal, and Jitendra Malik.
\newblock A database of human segmented natural images and its application to evaluating segmentation algorithms and measuring ecological statistics.
\newblock In \emph{Proceedings eighth IEEE international conference on computer vision. ICCV 2001}, pages 416--423. IEEE, 2001.

\bibitem[Menon et~al.(2020)Menon, Damian, Hu, Ravi, and Rudin]{pulse}
Sachit Menon, Alexandru Damian, Shijia Hu, Nikhil Ravi, and Cynthia Rudin.
\newblock Pulse: Self-supervised photo upsampling via latent space exploration of generative models.
\newblock In \emph{In CVPR}, pages 2437--2445, 2020.

\bibitem[Mittal et~al.(2012)Mittal, Soundararajan, and Bovik]{NIQE}
Anish Mittal, Rajiv Soundararajan, and Alan~C Bovik.
\newblock Making a “completely blind” image quality analyzer.
\newblock \emph{IEEE Signal processing letters}, 20\penalty0 (3):\penalty0 209--212, 2012.

\bibitem[Radford et~al.(2021)Radford, Kim, Hallacy, Ramesh, Goh, Agarwal, Sastry, Askell, Mishkin, Clark, et~al.]{clip}
Alec Radford, Jong~Wook Kim, Chris Hallacy, Aditya Ramesh, Gabriel Goh, Sandhini Agarwal, Girish Sastry, Amanda Askell, Pamela Mishkin, Jack Clark, et~al.
\newblock Learning transferable visual models from natural language supervision.
\newblock In \emph{International conference on machine learning}, pages 8748--8763. PMLR, 2021.

\bibitem[Ramesh et~al.(2021)Ramesh, Pavlov, Goh, Gray, Voss, Radford, Chen, and Sutskever]{dell_e}
Aditya Ramesh, Mikhail Pavlov, Gabriel Goh, Scott Gray, Chelsea Voss, Alec Radford, Mark Chen, and Ilya Sutskever.
\newblock Zero-shot text-to-image generation.
\newblock In \emph{International conference on machine learning}, pages 8821--8831. Pmlr, 2021.

\bibitem[Rombach et~al.(2022)Rombach, Blattmann, Lorenz, Esser, and Ommer]{stablediffusion}
Robin Rombach, Andreas Blattmann, Dominik Lorenz, Patrick Esser, and Bj{\"o}rn Ommer.
\newblock High-resolution image synthesis with latent diffusion models.
\newblock In \emph{In CVPR}, pages 10684--10695, 2022.

\bibitem[Saharia et~al.(2022{\natexlab{a}})Saharia, Chan, Saxena, Li, Whang, Denton, Ghasemipour, Gontijo~Lopes, Karagol~Ayan, Salimans, et~al.]{imagen}
Chitwan Saharia, William Chan, Saurabh Saxena, Lala Li, Jay Whang, Emily~L Denton, Kamyar Ghasemipour, Raphael Gontijo~Lopes, Burcu Karagol~Ayan, Tim Salimans, et~al.
\newblock Photorealistic text-to-image diffusion models with deep language understanding.
\newblock \emph{Advances in neural information processing systems}, 35:\penalty0 36479--36494, 2022{\natexlab{a}}.

\bibitem[Saharia et~al.(2022{\natexlab{b}})Saharia, Ho, Chan, Salimans, Fleet, and Norouzi]{sr3}
Chitwan Saharia, Jonathan Ho, William Chan, Tim Salimans, David~J Fleet, and Mohammad Norouzi.
\newblock Image super-resolution via iterative refinement.
\newblock \emph{IEEE transactions on pattern analysis and machine intelligence}, 45\penalty0 (4):\penalty0 4713--4726, 2022{\natexlab{b}}.

\bibitem[Schulman et~al.(2017)Schulman, Wolski, Dhariwal, Radford, and Klimov]{ppo}
John Schulman, Filip Wolski, Prafulla Dhariwal, Alec Radford, and Oleg Klimov.
\newblock Proximal policy optimization algorithms.
\newblock \emph{arXiv preprint arXiv:1707.06347}, 2017.

\bibitem[Song et~al.(2020)Song, Meng, and Ermon]{ddim}
Jiaming Song, Chenlin Meng, and Stefano Ermon.
\newblock Denoising diffusion implicit models.
\newblock \emph{arXiv preprint arXiv:2010.02502}, 2020.

\bibitem[Timofte et~al.(2017)Timofte, Agustsson, Van~Gool, Yang, and Zhang]{2017ntire}
Radu Timofte, Eirikur Agustsson, Luc Van~Gool, Ming-Hsuan Yang, and Lei Zhang.
\newblock Ntire 2017 challenge on single image super-resolution: Methods and results.
\newblock In \emph{Proceedings of the IEEE conference on computer vision and pattern recognition workshops}, pages 114--125, 2017.

\bibitem[Tsao et~al.(2024)Tsao, Lo, Chang, Chen, Tseng, Feng, and Lee]{tsao2024boosting}
Li-Yuan Tsao, Yi-Chen Lo, Chia-Che Chang, Hao-Wei Chen, Roy Tseng, Chien Feng, and Chun-Yi Lee.
\newblock Boosting flow-based generative super-resolution models via learned prior.
\newblock In \emph{Proceedings of the IEEE/CVF Conference on Computer Vision and Pattern Recognition}, pages 26005--26015, 2024.

\bibitem[Wang et~al.(2021{\natexlab{a}})Wang, Zhang, Yuan, and Wang]{wang2021unsupervised}
Wei Wang, Haochen Zhang, Zehuan Yuan, and Changhu Wang.
\newblock Unsupervised real-world super-resolution: A domain adaptation perspective.
\newblock In \emph{In ICCV}, pages 4318--4327, 2021{\natexlab{a}}.

\bibitem[Wang et~al.(2021{\natexlab{b}})Wang, Xie, Dong, and Shan]{wang2021real}
Xintao Wang, Liangbin Xie, Chao Dong, and Ying Shan.
\newblock Real-esrgan: Training real-world blind super-resolution with pure synthetic data.
\newblock In \emph{In ICCV}, pages 1905--1914, 2021{\natexlab{b}}.

\bibitem[Wang et~al.(2023)Wang, Chen, Ni, Wang, Tong, and Liu]{eqsr}
Xiaohang Wang, Xuanhong Chen, Bingbing Ni, Hang Wang, Zhengyan Tong, and Yutian Liu.
\newblock Deep arbitrary-scale image super-resolution via scale-equivariance pursuit.
\newblock In \emph{In CVPR}, pages 1786--1795, 2023.

\bibitem[Wei and Zhang(2023)]{srno}
Min Wei and Xuesong Zhang.
\newblock Super-resolution neural operator.
\newblock In \emph{In CVPR}, pages 18247--18256, 2023.

\bibitem[Wu et~al.(2024)Wu, Yang, Sun, Zhang, Li, and Zhang]{seesr}
Rongyuan Wu, Tao Yang, Lingchen Sun, Zhengqiang Zhang, Shuai Li, and Lei Zhang.
\newblock Seesr: Towards semantics-aware real-world image super-resolution.
\newblock In \emph{In CVPR}, pages 25456--25467, 2024.

\bibitem[Yang et~al.(2024)Yang, Wu, Ren, Xie, and Zhang]{pasd}
Tao Yang, Rongyuan Wu, Peiran Ren, Xuansong Xie, and Lei Zhang.
\newblock Pixel-aware stable diffusion for realistic image super-resolution and personalized stylization.
\newblock In \emph{European Conference on Computer Vision}. Springer, 2024.

\bibitem[Yao et~al.(2023)Yao, Tsao, Lo, Tseng, Chang, and Lee]{linf}
Jie-En Yao, Li-Yuan Tsao, Yi-Chen Lo, Roy Tseng, Chia-Che Chang, and Chun-Yi Lee.
\newblock Local implicit normalizing flow for arbitrary-scale image super-resolution.
\newblock In \emph{In CVPR}, pages 1776--1785, 2023.

\bibitem[Zeyde et~al.(2012)Zeyde, Elad, and Protter]{Set14}
Roman Zeyde, Michael Elad, and Matan Protter.
\newblock On single image scale-up using sparse-representations.
\newblock In \emph{Curves and Surfaces: 7th International Conference, Avignon, France, June 24-30, 2010, Revised Selected Papers 7}, pages 711--730. Springer, 2012.

\bibitem[Zhang et~al.(2021)Zhang, Liang, Van~Gool, and Timofte]{realsrset}
Kai Zhang, Jingyun Liang, Luc Van~Gool, and Radu Timofte.
\newblock Designing a practical degradation model for deep blind image super-resolution.
\newblock In \emph{In ICCV}, pages 4791--4800, 2021.

\bibitem[Zhang et~al.(2023)Zhang, Rao, and Agrawala]{contorlnet}
Lvmin Zhang, Anyi Rao, and Maneesh Agrawala.
\newblock Adding conditional control to text-to-image diffusion models.
\newblock In \emph{In ICCV}, pages 3836--3847, 2023.

\bibitem[Zhang et~al.(2018{\natexlab{a}})Zhang, Isola, Efros, Shechtman, and Wang]{lpips}
Richard Zhang, Phillip Isola, Alexei~A Efros, Eli Shechtman, and Oliver Wang.
\newblock The unreasonable effectiveness of deep features as a perceptual metric.
\newblock In \emph{Proceedings of the IEEE conference on computer vision and pattern recognition}, pages 586--595, 2018{\natexlab{a}}.

\bibitem[Zhang et~al.(2018{\natexlab{b}})Zhang, Li, Li, Wang, Zhong, and Fu]{rcan2018}
Yulun Zhang, Kunpeng Li, Kai Li, Lichen Wang, Bineng Zhong, and Yun Fu.
\newblock Image super-resolution using very deep residual channel attention networks.
\newblock In \emph{Proceedings of the European conference on computer vision (ECCV)}, pages 286--301, 2018{\natexlab{b}}.

\bibitem[Zhang et~al.(2024)Zhang, Huang, Ma, Li, Luo, Xie, Qin, Luo, Li, Liu, et~al.]{ram}
Youcai Zhang, Xinyu Huang, Jinyu Ma, Zhaoyang Li, Zhaochuan Luo, Yanchun Xie, Yuzhuo Qin, Tong Luo, Yaqian Li, Shilong Liu, et~al.
\newblock Recognize anything: A strong image tagging model.
\newblock In \emph{In CVPR}, pages 1724--1732, 2024.

\bibitem[Zheng et~al.(2018)Zheng, Jiang, and Huang]{strokenet}
Ningyuan Zheng, Yifan Jiang, and Dingjiang Huang.
\newblock Strokenet: A neural painting environment.
\newblock In \emph{International Conference on Learning Representations}, 2018.

\bibitem[Zou et~al.(2021)Zou, Shi, Qiu, Yuan, and Shi]{stylized}
Zhengxia Zou, Tianyang Shi, Shuang Qiu, Yi Yuan, and Zhenwei Shi.
\newblock Stylized neural painting.
\newblock In \emph{In CVPR}, pages 15689--15698, 2021.

\end{thebibliography}
}


\end{document}